\documentclass[10pt,twocolumn,letterpaper]{article}

\usepackage{iccv}              

\definecolor{iccvblue}{rgb}{0.21,0.49,0.74}
\usepackage[pagebackref,breaklinks,colorlinks,allcolors=iccvblue]{hyperref}
\usepackage{pifont}  
\usepackage{multirow}
\usepackage{booktabs}
\usepackage{colortbl}
\usepackage[normalem]{ulem} 
\usepackage{soul}
\usepackage{tikz}

\definecolor{lightgreen}{RGB}{144, 238, 144} 
\definecolor{darkgreen}{RGB}{0, 100, 0}     
\definecolor{mypurple}{RGB}{128, 0, 128}    

\def\paperID{2013} 
\def\confName{ICCV}
\def\confYear{2025}

\title{Implicit Counterfactual Learning for Audio-Visual Segmentation}

\author{Mingfeng Zha$^{1}$, Tianyu Li$^{1}$\thanks{Corresponding author.}, Guoqing Wang$^{1}$, Peng Wang$^{1}$, Yangyang Wu$^{2}$, Yang Yang$^{1}$,  Heng Tao Shen$^{3}$\\
$^{1}$University of Electronic Science and Technology of China, $^{2}$Zhejiang University, $^{3}$Tongji University\\
}

\usepackage{caption}
\begin{document}
\maketitle
\begin{abstract}
Audio-visual segmentation (AVS) aims to segment objects in videos based on audio cues. Existing AVS methods are primarily designed to enhance interaction efficiency but pay limited attention to modality representation discrepancies and imbalances. To overcome this, we propose the implicit counterfactual framework (ICF) to achieve unbiased cross-modal understanding. Due to the lack of semantics, heterogeneous representations may lead to erroneous matches, especially in complex scenes with ambiguous visual content or interference from multiple audio sources. We introduce the multi-granularity implicit text (MIT) involving video-, segment- and frame-level as the bridge to establish the modality-shared space, reducing modality gaps and providing prior guidance. Visual content carries more information and typically dominates, thereby marginalizing audio features in the decision-making. To mitigate knowledge preference, we propose the semantic counterfactual (SC) to learn orthogonal representations in the latent space, generating diverse counterfactual samples, thus avoiding biases introduced by complex functional designs and explicit modifications of text structures or attributes. We further formulate the collaborative distribution-aware contrastive learning (CDCL), incorporating factual-counterfactual and inter-modality contrasts to align representations, promoting cohesion and decoupling. Extensive experiments on three public datasets validate that the proposed method achieves state-of-the-art performance. 
\end{abstract}    
\section{Introduction}
\label{sec:intro}

Referring video segmentation \cite{he2024decoupling,yuan2024losh,zhu2024exploring,yan2024referred} which utilizes text or audio prompts to identify matching targets in visual content, has been applied to diverse tasks such as embodied intelligence \cite{gupta2021embodied} and autonomous driving \cite{wang2024drivedreamer}. Unlike highly structured text, the low density and fuzziness of audio align more closely with natural attributes, posing significant challenges for audio-visual segmentation (AVS).

\begin{figure*}[tb]
  \centering
  \includegraphics[width=1\textwidth]{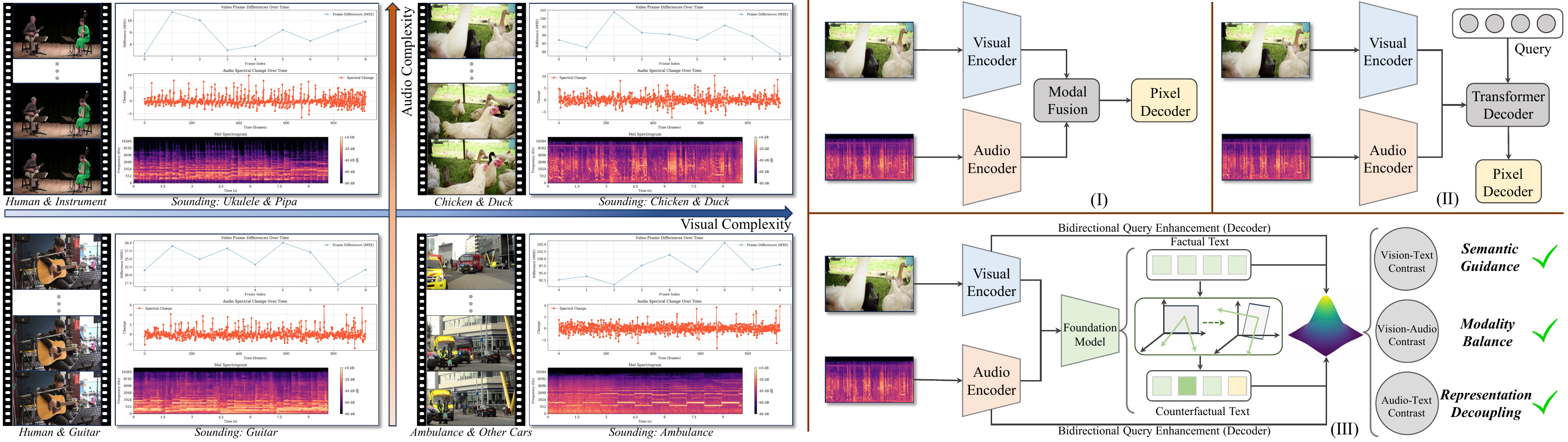}
  \caption{The problem decomposition and method paradigm of AVS task. In the left subfigure, we divide the task into four quadrants based on two metrics: Mean Squared Error (MSE) to measure inter-frame content changes (reflecting visual complexity) and spectral changes (Mel spectrogram) to characterize audio complexity. The dependence of visual complexity on audio exhibits a certain positive correlation. In the right subfigure, existing works can be categorized into (\text{I}) Modal fusion; (\text{II}) Query learning (with numerous variants prompt-based strategies that are challenging to unify). Our approach (\text{III}) involves counterfactual implicit text generation and Gaussian spatial modality contrast, which can be adapted to paradigm (\text{I}) and paradigm (\text{II}).}
  \label{Intro}
\end{figure*}

In Figure \ref{Intro}, we categorize the AVS issues into four parts based on the complexity of audio and visual inputs. From left to right, the sounding targets transition from static to highly dynamic scenarios, ranging from cases where visual segmentation models alone suffice to those requiring robust temporal and cross-modal joint representations. From bottom to top, as the audio sources evolve from single-source to multi-source and multi-class, the need for decoupling modality representations and achieving accurate alignment becomes increasingly critical. These challenges can be summarized as: \textit{a) Multi-source audio and complex dynamics}, where the simultaneous presence of multiple audio sources and temporal variations complicates effective association, resulting in mismatches, particularly in complex scenes within frames and rapid changes across frames, causing low intra-modal cohesion; \textit{b) Learning preference}, where models favor high information-dense visual features and underutilize sparse audio, leading to weak inter-modal coupling. Current works \cite{hao2024improving,liu2023audio,gao2024avsegformer,chen2024bootstrapping} primarily design problem-specific functional components and combining them. Although effective, learning optimal parameters in a fixed data space risks capturing spurious correlations. For instance, since guitars are typically played by humans to produce sound, models might incorrectly associate humans as sound producers based on statistical regularities.  Inspired by the human brain's ability to self-project and simulate unreal scenarios \cite{buckner2007self,addis2007remembering,coricelli2005regret}, we formulate the ICF, which leverages text to bridge visual and audio modalities, constructs counterfactual text samples, and establishes the contrastive strategy based on representation distributions. This raises three key questions: \textit{1) Why use language as a bridge instead of direct interaction? 2) Why construct implicit counterfactual samples? 3) Why construct contrast learning based on representation distribution?}

\textit{We answer the first question.} In complex scenes, multiple visual objects may emit sounds synchronously or asynchronously (many-to-many), or a single sound may correspond to multiple visual objects (one-to-many), making direct matching prone to ambiguities. Furthermore, temporal delays between visual content and audio can exacerbate the issue. We aim to leverage the semantic information of text to capture the objects, actions, and their relationships within the scene, facilitating cross-modal alignment and modeling contextual continuity across frames (\ul{\textbf{\textit{challenge a}}}). Differing from \cite{wang2024can,liu2024bavs}, which generates explicit texts/categories for visual end, we match the best implicit texts at the feature level for both vision and audio. This approach is based on: a) Implicit texts, derived from higher-order spaces, are less susceptible to low-level noise and more suitable for constructing counterfactual samples; b) Unlike visual semantics, which often encompass silent objects, audio semantics are inherently localized and focused.

\textit{We answer the second question.} To alleviate the knowledge discrepancies in data structure, some works \cite{yu2024revisiting,lai2024improving} modify explicit text attributes, such as nouns, colors, sizes, or spatial orientations, or employ generative models (\textit{e.g.,} DALL-E \cite{ramesh2021zero}) to create counterfactual images or adjust the audio spectrum (\ul{\textbf{\textit{challenge b}}}). However, explicit strategies typically require selecting elements for modification from the predefined pool of assumptions, with the upper bound of possibilities potentially smaller than implicit strategies. Our approach also ensures that representations of factual and counterfactual texts progressively diverge, facilitating curriculum-based model training. Additionally, explicit strategies incur significant storage and training costs and are challenging to implement in an end-to-end manner.

\textit{We answer the third question.} Previous audio-visual works \cite{chen2024unraveling,chen2024cpm} construct contrastive learning at the feature level, which is sensitive to internal structural variations. In contrast, our strategy focuses on: a) Leveraging statistical distributions to model contextual features, enhancing robustness to scene changes, audio noises/mixtures, and inaccuracies in implicit texts, thereby mitigating incorrect matches caused by hard contrastive learning; b) Ensuring stable alignment between visual and audio features with factual texts while maintaining internal compactness, thus preventing degradation from counterfactual texts. 

Technically, we propose the multi-granularity implicit text (MIT), which takes video-, segment-, and frame-level visual features as inputs to the foundation models to retrieve best matching implicit textual representations. The similar principles are applied to audio features. We further introduce the semantic counterfactual (SC), which leverages the latent diffusion model to establish continuous and controllable intra-sample and inter-sample orthogonality in the noise space during the forward process, generating counterfactual samples through the denoising process. Finally, we formulate the collaborative distribution-aware contrastive learning (CDCL), which transforms modality features into Gaussian distributions and constructs contrasts in the joint embedding space based on an auxiliary entropy (uncertainty) metric. Our main contributions are as follows:
\begin{itemize}
\setlength{\itemsep}{0pt}
\setlength{\parsep}{0pt}
\setlength{\parskip}{0pt}
\item To the best of our knowledge, we are the first to model learning biases in the AVS task based on causal inference theory and introduce implicit text to generate counterfactual samples, achieving semantic guidance and balance.
\item We propose the MIT and the SC to establish semantic correlations and variances in the continuous space, and further formulate the CDCL to decouple and cohere modality feature distributions.
\item Extensive experiments on three public datasets demonstrate that our method achieves state-of-the-art results and can be seamlessly integrated into other AVS approaches, improving performance by 3\%-4\%.
\end{itemize}
\section{Related Work}
\label{sec:related work}

\noindent\textbf{Audio-Visual Segmentation.} Audio-visual source localization (AVL) \cite{mahmud2024t,kim2024learning,mo2023audio,sun2023learning,hu2022mix} determines the approximate location of sound-emitting objects in videos by leveraging audio cues at the regional level, primarily through unsupervised learning to establish cross-modal representation correlations. Recently, Zhou \textit{et al.} \cite{zhou2022audio} advanced the AVL task by exploring pixel-level audio-visual understanding and introducing the AVS task. Existing fully supervised AVS methods can be broadly categorized into feature decoupling and fusion-based \cite{yang2024cooperation,li2024qdformer,chen2024unraveling,liu2024audio,liu2024benchmarking,mao2023multimodal,ma2024stepping,hao2024improving,li2024selm}, query generation-based \cite{gao2024avsegformer,huang2023discovering,li2023catr,liu2023audio}, and prompt injection-based \cite{chen2024cpm,sun2024unveiling,wang2024prompting,liu2023annotation,liu2024bavs}. Additionally, some studies further explored efficient label learning \cite{mo2023weakly,bhosale2024unsupervised,guo2024open}, 3D spatial perception \cite{sokolov20243d}, and medical scenario \cite{chen2024asi}. However, the above works generally overlook biases in learning caused by differences in modality representations and distributions. Unlike \cite{sun2024unveiling}, which mitigates model preferences via active queries and biased branches, we aim to achieve semantic guidance and representation balance through implicit texts and counterfactual samples.

\noindent\textbf{Contrastive Learning.} Contrastive learning \cite{zha2025heterogeneous,zha2024weakly} aims to minimize the distance between paired samples/categories while maximizing the distance between unmatched samples/different categories. \cite{chen2024cpm,chen2024unraveling,mao2023contrastive,wang2023prompting} constructed positive and negative sample pairs using semantic categories (prompts) as anchors to facilitate representation disentanglement. In contrast, \cite{sun2024unveiling} focused exclusively on audio features. Our strategy differs in two aspects: 1) Measuring modality differences from representation distributions; 2) Establishing heterogeneous tri-modal contrastive learning.

\noindent\textbf{Counterfactual Learning.} Based on the ladder theory \cite{pearl2018book} of 
causal inference, previous AVS works focused on constructing biased modality associations (\textit{i.e.,} \textit{if..., then...}). Due to the interference of confounding factors, statistically strong correlations between two variables do not necessarily imply causality. Intervention and counterfactual serve as effective methods to achieve impartiality and have been applied in various scenarios, \textit{e.g.,} visual question answering \cite{zang2023discovering,liu2023cross}, image segmentation \cite{zhang2020causal,chen2022c}, and vision-language navigation \cite{wang2022counterfactual,wang2024vision}. Exploring unexperienced situations through intervention poses significant challenges. We instead leverage counterfactual reasoning (\textit{i.e.,} \textit{if not..., then...}) to construct the continuous hypothesis space, thereby expanding the potential boundaries of samples and establishing unbiased and accurate correlations.

\section{Methodology}
\label{sec:method}

\subsection{Overview}

\begin{figure*}[tb]
  \centering
  \includegraphics[width=1\textwidth]{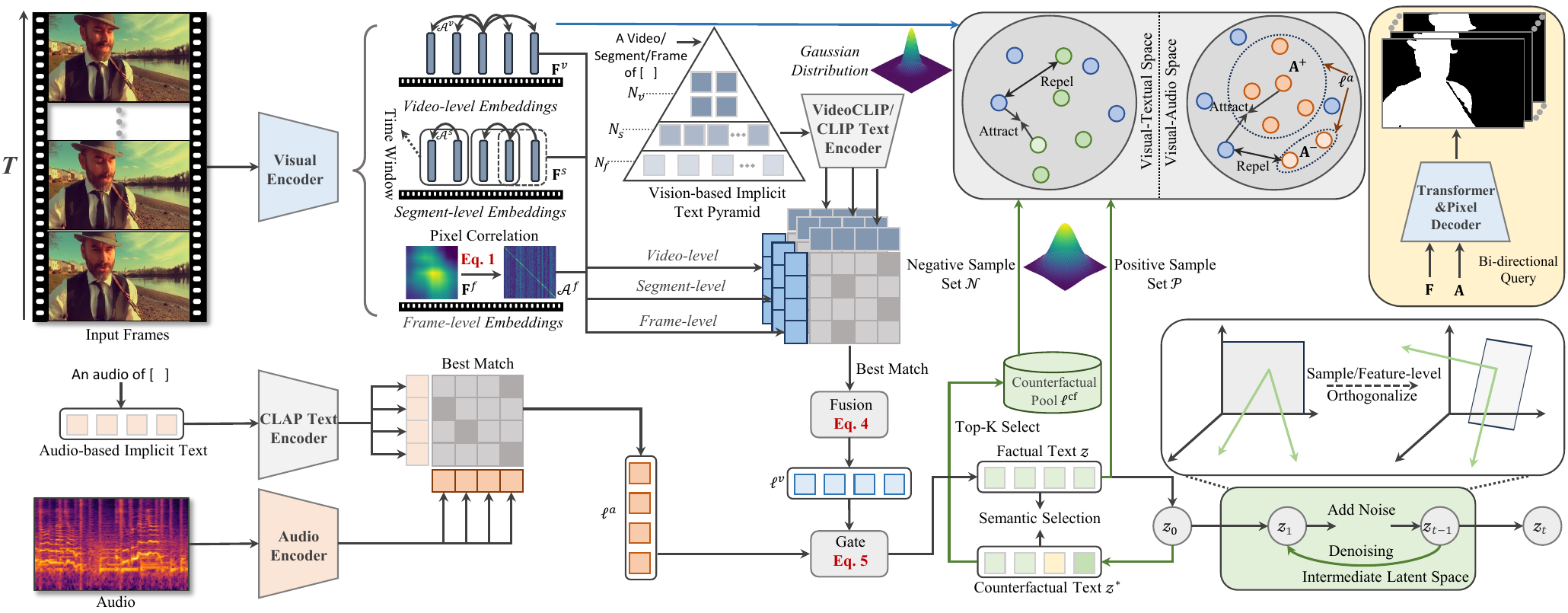}
  \caption{The framework of ICF. For any given audio-video pair, we obtain high-dimensional features using visual and audio encoders separately. For visual features, we establish global temporal correlations, clip semantic partitioning, and intra-frame context alignment at video-segment-frame levels. Subsequently, we initialize a visual-conditioned implicit text pyramid and search for the best elements that match the semantics and adaptively fuse them. For audio features, we perform similar operations using CLAP. Through gate filtering, we generate counterfactual text pools by orthogonalizing factual texts in the diffusion latent space. We transform modality features into distributions and form contrastive learning. Bi-directional query learning \cite{gao2024avsegformer,chen2024bootstrapping} serves as the foundation for subsequent decoding.}
  \label{Framework}
\end{figure*}

In Figure \ref{Framework}, we utilize the visual encoder to generate multi-scale features $\mathbf{F} \in \mathbb{R}^{T \times C_{i} \times H_{i} \times W_{i}}$ ($T$, $C$, $H$, $W$ denote the number of frames, channels, width, and height, respectively). For audio, we perform short-time Fourier transform to obtain Mel-spectrograms, which are then processed by the audio encoder to generate audio features $\mathbf{A} \in \mathbb{R}^{T \times D}$ ($D$ denotes feature dimension). For visual features, we leverage foundation models to match text features corresponding to video-, segment- and frame-level  features, \textit{i.e.,} implicit text $\ell^{v}$. Similar operations are conducted on audio features to generate $\ell^{a}$, further forming the composite factual text $z$. Subsequently, we add noise to $z$ based on the latent diffusion model and orthogonalize partial representations in the noise space, generating counterfactual samples $\ell^{\rm cf}$ via the denoising process. Finally, we model the probability space of tri-modal features based on Gaussian distributions and conduct contrastive learning across paired modalities.

\subsection{Multi-granularity Implicit Text}
Inspired by \cite{heekeren2004general}, humans associate visual content with auditory cues, typically relying on: 1) Global scene perception (video-level); 2) Inter-frame object motion and semantic change (segment-level); 3) Current moment context (frame-level). For video-level features, we construct long-range pixel-level dependencies between frames. Specifically, we utilize three matrices to map $\mathbf{F}^{v}$ into query $\mathbf{Q}^{v}$, key $\mathbf{K}^{v}$, and value $\mathbf{V}^{v}$, thereby obtaining the correlation matrix $\mathcal{A}^{v}$,
\begin{equation}
\setlength\abovedisplayskip{3pt}
\setlength\belowdisplayskip{3pt}
\mathcal{A}^{v} = \texttt{Softmax} \left( \frac{\mathbf{Q}^v (\mathbf{K}^v)^{\top}}{\tau}\right) \in \mathbb{R}^{TC \times TC}
\end{equation}
where $\tau$ and $\top$ represent the temperature coefficient and transpose, respectively. To reduce computational costs, we establish spatiotemporal correlations from the channel dimension rather than the spatial dimension (complexity is reduced from $\mathcal{O}(H^2W^2)$ to $\mathcal{O}(C^2)$). We further update,
\begin{equation}
\setlength\abovedisplayskip{3pt}
\setlength\belowdisplayskip{3pt}
\mathbf{F}^v := \mathbf{F}^v + \mathbf{V}^v\mathcal{A}^v
\end{equation}

For segment-level features, we employ similar strategy for sampling. The difference lies in setting the time window to $\lceil \frac{T}{2} \rceil$ or $\lceil \frac{T}{4
} \rceil$ and sliding it backward. When the number of frames within the window is insufficient, we utilize preceding frames for supplementation. For frame-level features, the distinction lies in the transition from inter-frame temporal to intra-frame context correlation. Based on textual inversion \cite{gal2022image,tan2023language}, we utilize VideoCLIP \cite{xu2021videoclip} to generate implicit text $l^{v}$ corresponding to $\mathbf{F}^v$,
\begin{equation}
\setlength\abovedisplayskip{3pt}
\setlength\belowdisplayskip{3pt}
l^{v} \in \arg\max_{l_1,l_2,\ldots,l_k}\frac{1}{k^t}\sum_{i=1}^{k^t} \texttt{sim} \left(\mathbf{F}^{v}, \texttt{VideoCLIP}(l_i)\right)
\end{equation}
where we introduce $k^t$ learnable parameters $l$ to capture multiple visual concepts to avoid focusing exclusively on the most salient semantics. For $\mathbf{F}^s$, likewise. For each frame feature $\mathbf{F}^{f}$, we obtain $T$ factual texts using CLIP \cite{radford2021learning}. Thus, we can obtain semantic representations ranging from coarse to fine granularity. To measure the contribution of each element, we introduce a weight query $w$ to dynamically adjust the weights, forming the fused text features $\ell^{v}$,
\begin{equation}
\setlength\abovedisplayskip{3pt}
\setlength\belowdisplayskip{3pt}
\ell^{v} = \sum_{p \in \{v, s, f\}} \sum_{n=1}^{N_p}\left( \frac{\exp({w}^{p}_{n})}{\sum_{q\in\{v,s,f\}}\sum_{m=1}^{N_q} \exp({w}^{q}_{m})} \cdot {l}^{p}_{n} \right)
\end{equation}
where $N_p$ represents the number of factual texts corresponding to each scale. For audio features, we use CLAP \cite{wu2023large} to match the audio factual text $\ell^{a}$. We introduce gate mechanism (three MLP \cite{liu2021pay} layers) to alleviate semantics conflicts, and further generate composite factual text $z$,
\begin{equation}
\setlength\abovedisplayskip{3pt}
\setlength\belowdisplayskip{3pt}
z = \mathcal{F}(\texttt{Concat}(\texttt{Gate}(\ell^{v}), \texttt{Gate}(\ell^{a})))
\end{equation}
where $\texttt{Concat}$ denotes dimension-wise concatenation. 

\subsection{Semantic Counterfactual}
At the feature-level, counterfactual samples can be generated directly by adding noise or applying transformations. This strategy may disrupt the original semantic distribution and be uncontrollable (Table \ref{Quantitative ablation of proposed components}). We utilize the stable sample construction capability of diffusion model to impose orthogonality in the latent space. To prevent learning direction confusion, we condition on visual features $v$ and gradually add Gaussian noise to $z$ in the forward process,
\begin{equation}
\setlength\abovedisplayskip{3pt}
\setlength\belowdisplayskip{3pt}
q(z_{t} | z_{t-1},v) = \mathcal{N}(z_{t}; \sqrt{1 - \beta_{t}} z_{t-1}, \beta_{t} I)
\end{equation}
where $\beta$, $I$ denote variance schedule parameter and identity matrix, respectively. We normalize $z$ and then orthogonalize based on the Gram-Schmidt strategy \cite{bjorck1994numerics},
\begin{equation}
\setlength\abovedisplayskip{3pt}
\setlength\belowdisplayskip{3pt}
r_{(i)} = \| r_{(i)} - (r_{(i)} \cdot z_{(i)})z_{(i)} \|, \ r \sim \mathcal{N}(0,I)
\end{equation}
where $r$ is a random matrix with the same dimension as $z$. To control the degree of counterfactual, we introduce $\alpha$ to adjust the ratio. However, controlling all samples within the same batch to maintain the same degree does not align with reality. We further introduce $m, s \in (0,1]$ to ensure the inter-sample and intra-sample feature variations,
\begin{equation}
\begin{split}
\setlength\abovedisplayskip{3pt}
\setlength\belowdisplayskip{3pt}
z_{(i)}' = \sqrt{1 - \alpha \cdot m_{(i)}} \cdot z_{(i)} + \sqrt{\alpha \cdot m_{(i)}} \cdot r_{(i)}, \\
z_{(j)}' = \sqrt{1 - \alpha \cdot s_{(j)}} \cdot z_{(j)} + \sqrt{\alpha \cdot s_{(j)}} \cdot r_{(j)}
\end{split}
\end{equation}
when $\alpha \cdot m\ (\texttt{or}\ s) =1$, completely orthogonality is achieved; when $\alpha \cdot m \ (\texttt{or}\ s) =0$, $z$ remains unchanged. In this way, we can customize the heterogeneity for each sample. During the reverse process, we gradually denoise to generate counterfactual text $z^{*}$ conditioned on $z^{\prime}$.
\begin{equation}
\setlength\abovedisplayskip{3pt}
\setlength\belowdisplayskip{3pt}
p_{\theta}\big(z^{\prime}_{t-1} \mid z^{\prime}_{t}, v\big) = \mathcal{N}\big(z^{\prime}_{t-1}; \mu_{\theta}(z^{\prime}_{t}, t, v), \Sigma_{\theta}(z^{\prime}_{t}, t, v)\big)
\end{equation}
where $t$, $\mu_{\theta}$ $\Sigma_{\theta}$ respectively represent time step, the mean, and covariance prediction functions based on the model parameters $\theta$. For control parameters, we analyze two options: \textit{1) Hyperparameters}, fixed values that require manual design and are difficult to dynamically adjust based on video content; \textit{2) Learnable parameters}, although capable of change through learning, may converge to local optima. Combining the two, we establish the boundary for learning. We aim to constrain the optimization direction through loss,
\begin{equation}
\setlength\abovedisplayskip{3pt}
\setlength\belowdisplayskip{3pt}
\mathcal{L}_{\rm ortho} = \|z^{\prime}_{t} - z_{t}\|^2 + \lambda_{z} \|z^{\prime}_{t} \cdot z_{t}\|^2
\end{equation}
where $\lambda_{z}$ is the balancing parameter. The first and second terms respectively ensure that the features maintain a certain level of similarity and orthogonality. $z^{\prime}$ is derived from $z$ through the orthogonal transformation, preserving the internal noise structure. In other words, the noises added and removed come from the same distribution, \textit{i.e.,} noise symmetry. We leverage the objective function in \cite{rombach2022high} to estimate the noise. The counterfactual text generation loss $\mathcal{L}_{\rm cf}$ can be formulated as,
\begin{equation}
\setlength\abovedisplayskip{3pt}
\setlength\belowdisplayskip{3pt}
\mathcal{L}_{\rm cf} = \mathbb{E}_{t, z_{t}, \epsilon} \left[ \left\| \epsilon - \epsilon_{\theta}(z^{\prime}_{t}, t) \right\|^{2} \right] + \lambda_{\rm ortho} \mathcal{L}_{\rm ortho}
\end{equation} 
where $\epsilon$ and $\epsilon_{\theta}$ represent the actual added noise and the predicted noise, respectively. We select the Top-K samples that are semantically close in a continuous manner to form the counterfactual text pool,
\begin{equation} 
\setlength\abovedisplayskip{3pt}
\setlength\belowdisplayskip{3pt}
\ell^{\rm cf} = \arg\,\mathop{\texttt{TopK}}\limits_{z^{*}_1,z^{*}_2,\ldots,z^{*}_k} (\texttt{sim} (z^{*}, z))
\end{equation}
where $\texttt{sim}$ denotes cosine similarity. For samples with significant semantic differences, \textit{i.e.,} simple ones, they occupy much storage space but exhibit the marginal effect for subsequent contrasts.

\noindent \textbf{Discussion.} The SC operates based on the MIT, and both are only applied during the training phase. {\ul{\textit{In the fully latent space (pure noise), orthogonality is ineffective, thus counterfactual generation intervenes in the intermediate state.}}}

\subsection{Collaborative Distribution-aware Contrastive Learning}
Feature-level contrastive learning is sensitive to anomalies, \textit{e.g.,} blank video frames, mixed scenes, and drastic changes in audio spectra, which may lead to erroneous sample attraction and repulsion. This issue is further exacerbated when implicit textual semantic descriptions of video or audio content are inaccurate. To overcome this, we transform features into statistical probabilities to enhance tolerance towards anomalies. Specifically, we model the $T$ frames using Gaussian distribution to potentially aggregate sequential information, generating mean ($\mu^{v}$) and covariance ($\Sigma^{v}$),
\begin{equation}
\setlength\abovedisplayskip{3pt}
\setlength\belowdisplayskip{3pt}
\mu^{v} = \frac{1}{T} \sum_{t=1}^{T} \mathbf{F}_{t}^{v}, \ \Sigma^{v} = \frac{1}{T} \sum_{t=1}^{T} (\mathbf{F}_{t}^{v} - \mu^{v})(\mathbf{F}_{t}^{v} - \mu^{v})^{\top}
\label{statistical}
\end{equation}

The introduction of low-quality data \cite{chen2024unraveling} increases aleatoric uncertainty \cite{hora1996aleatory}. Prior works pay little attention to this and treat features from different modalities equally. Although we alleviate through distribution modeling, accurate quantification remains essential. We utilize continuous information entropy \cite{shannon1948mathematical} $\mathcal{H}$ to reflect the average uncertainty,
\begin{equation}
\setlength\abovedisplayskip{3pt}
\setlength\belowdisplayskip{3pt}
{\cal H}(v) = \frac{1}{2} \log \left( (2 \pi e)^{d} \, \texttt{det}(\Sigma^{v}) \right)
\label{entropy}
\end{equation}
where $d$ and $\texttt{det}$ denote the dimension of $v$ and the determinant of the matrix, respectively. Similarly, we incorporate audio features $a$ into Eq. \ref{statistical} and Eq. \ref{entropy} to obtain the corresponding statistical metrics. We adopt Wasserstein Distance \cite{ruschendorf1985wasserstein} along with modality entropy to measure the distance between distributions,
\begin{equation}
\small
\setlength\abovedisplayskip{3pt}
\setlength\belowdisplayskip{3pt}
\begin{split}
    \mathcal{D}(v, a) & = \left\|\mu^{v} - \mu^{a}\right\|_{2}^{2} + {\texttt{Tr}}\left(\Sigma^{v} + \Sigma^{a} - 2(\Sigma^{v}\Sigma^{a})^{1/2}\right) \\
    & \quad + \gamma (\mathcal{H}(v)+\mathcal{H}(a))
\end{split}
\end{equation}
where $\texttt{Tr}$ denotes the trace of the matrix. The latter term measures the looseness (distance) within the distribution, where the low entropy (\textit{i.e.,} low uncertainty) of the joint modality distribution indicates high cohesion. To facilitate visual-audio contrast, the most straightforward approach, \textit{i.e.,} InfoNCE loss \cite{he2020momentum}, is to consider the same $\langle \mathbf{F}^v, \mathbf{A} \rangle$ pairs as the positive sample set $\mathcal{P}$ and the rest as the negative sample set $\mathcal{N}$. However, this hard contrast strategy may cause videos with the same audio semantics to repel each other. We partition the sample space using $\ell^a$ as the anchor (visual content variations make $z$ unreliable as the reference) to generate $\mathbf{A}^{+}$ and $\mathbf{A}^{-}$. We formulate $\mathcal{L}_{\rm v \leftrightarrow a}$ as, 
\begin{equation}
\small
\setlength\abovedisplayskip{3pt}
\setlength\belowdisplayskip{3pt}
\begin{split}
    & \mathcal{L}_{\rm v \leftrightarrow a} = -\frac{1}{B} \sum_{i=1}^{B} 
    \log \left[ \sum_{j \in \mathcal{P}(i)} \mathcal{D}^{\prime}(\mathbf{F}_i^v, \mathbf{A}^{+}_j)  \right. \\
    & \left. \frac{1}{\sum_{j \in \mathcal{P}(i)} \mathcal{D}^{\prime}(\mathbf{F}_i^v, \mathbf{A}^{+}_j)  + \sum_{k \in \mathcal{N}(i)} \mathcal{D}^{\prime}(\mathbf{F}_i^v, \mathbf{A}^{-}_k)} \right]
\end{split}
\end{equation}
where $\mathcal{D}^{\prime}=\exp(\mathcal{D}(\cdot,\cdot)/\tau)$, and $B$ represents the batch size. For visual-textual contrast, we define $\langle \mathbf{F}^v, z \rangle$ as the $\mathcal{P}$ and $\langle \mathbf{F}^v, \ell^{\rm cf}_{k} \rangle$ as the $\mathcal{N}$, without contrasting against other samples within the batch. Orthogonality determines the difference/similarity between counterfactual and factual samples. Thus, $k$-\textit{th} hard samples (low counterfactual) require assigning higher weights $w_k = \frac{\sqrt{1 - \alpha_k}}{\sum_{j=1}^{K} \sqrt{1 - \alpha_j}}$ (for simplicity, disregard $m$ and $s$). We formulate $\mathcal{L}_{\rm v \leftrightarrow l}$ as, 
\begin{equation}
\small
\setlength\abovedisplayskip{3pt}
\setlength\belowdisplayskip{3pt}
\mathcal{L}_{\rm v \leftrightarrow l} = -\frac{1}{B} \sum_{i=1}^{B} \log \left[ \frac{\mathcal{D}^{\prime}(\mathbf{F}_i^v, z_i)}{ \mathcal{D}^{\prime}(\mathbf{F}_i^v, z_i ) + \sum_{k \in \mathcal{N}(i)} w_k \cdot \mathcal{D}^{\prime}(\mathbf{F}_i^v, \ell^{\rm cf}_k) } \right]
\end{equation}

Similarly, we can formulate audio-text contrast $\mathcal{L}_{\rm a \leftrightarrow l}$.

\subsection{Loss Function}
Our main loss consists of mask segmentation $\mathcal{L}_{\rm Seg}$ and contrastive constraints $\mathcal{L}_{\rm contrast}$. We formulate  $\mathcal{L}_{\rm Seg}$ by,
\begin{equation}
\setlength\abovedisplayskip{3pt}
\setlength\belowdisplayskip{3pt}
\mathcal{L}_{\rm Seg} = \lambda_{\rm bce}\mathcal{L}_{\rm bce} + \lambda_{\rm dice}\mathcal{L}_{\rm dice} + \lambda_{\rm focal}\mathcal{L}_{\rm focal}
\end{equation}
where sub-items are binary cross-entropy, dice \cite{li2019dice}, and focal loss \cite{ross2017focal}. Thus, the total loss can be formulated by,
\begin{equation}
\small
\setlength\abovedisplayskip{3pt}
\setlength\belowdisplayskip{3pt}
\mathcal{L}_{\rm Total} = \mathcal{L}_{\rm Seg} + \lambda_{\rm cf} \mathcal{L}_{\rm cf} + \lambda_{\rm CDCL}\sum_{\rm p,\rm q \in \{\rm a,\rm v,\rm l\}}\lambda_{\rm p \leftrightarrow \rm q} \mathcal{L}_{\rm p \leftrightarrow q}
\end{equation}

\section{Experiments}
\label{sec:exp}

\begin{table*}[t]
\centering
\caption{Quantitative comparison of $\mathcal{J}$, $\mathcal{F}$ and $\mathcal{J}$\&$\mathcal{F}$ on the S4 \cite{zhou2022audio}, M3 \cite{zhou2022audio}, and AVSS \cite{zhou2024audio} datasets. External models, \textit{e.g.,} SAM \cite{kirillov2023segment}. $\uparrow$ represents the larger the better. Best performance in \textbf{bold}, second best \underline{underlined}. $\ddagger$ represents data is unavailable}
\resizebox{\textwidth}{!}{ 
\begin{tabular}{l|c|c|ccc|ccc|ccc}
\toprule
\multirow{2}{*}{Method} & \multirow{2}{*}{\begin{tabular}{@{}c@{}}External \\ Model\end{tabular}} & \multirow{2}{*}{Backbone} & \multicolumn{3}{c|}{AVS-Object (S4)} & \multicolumn{3}{c|}{AVS-Object (M3)} & \multicolumn{3}{c}{AVSS} \\
 &  &  & $\mathcal{J}\&\mathcal{F} \uparrow$ & $\mathcal{J} \uparrow$ & $\mathcal{F} \uparrow$ & $\mathcal{J}\&\mathcal{F} \uparrow$ & $\mathcal{J} \uparrow$ & $\mathcal{F} \uparrow$ & $\mathcal{J}\&\mathcal{F} \uparrow$ & $\mathcal{J} \uparrow$ & $\mathcal{F} \uparrow$ \\
\midrule
AVSBench \cite{zhou2022audio} \textsubscript{ECCV22} & \ding{55} & ResNet-50 & 78.80 & 72.79 & 84.80 & 52.84 & 47.88 & 57.80 & $\ddagger$ & $\ddagger$ & $\ddagger$ \\
ECMVAE \cite{mao2023multimodal} \textsubscript{ICCV23} & \ding{55} & ResNet-50 & 81.42 & 76.33 & 86.50 & 54.70 & 48.69 & 60.70 & $\ddagger$ & $\ddagger$ & $\ddagger$ \\
CATR \cite{li2023catr} \textsubscript{ACM MM23} & \ding{55} & ResNet-50 & 80.70 & 74.80 & 86.60 & 59.05 & 52.80 & 65.30 & $\ddagger$ & $\ddagger$ & $\ddagger$ \\
AQFormer \cite{huang2023discovering} \textsubscript{IJCAI23} & \ding{55} & ResNet-50 & 81.70 & 77.00 & 86.40 & 61.30 & 55.70 & 66.90 & $\ddagger$ & $\ddagger$ & $\ddagger$ \\
AVSegFormer \cite{gao2024avsegformer} \textsubscript{AAAI24} & \ding{55} & ResNet-50 & 80.67 & 76.54 & 84.80 & 56.17 & 49.53 & 62.80 & 27.12 & 24.93 & 29.30 \\
AVSC \cite{liu2023audio} \textsubscript{ACM MM23} & \ding{55} & ResNet-50 & 81.13 & 77.02 & 85.24 & 55.55 & 49.58 & 61.51 & $\ddagger$ & $\ddagger$ & $\ddagger$ \\
BAVS \cite{liu2024bavs} \textsubscript{TMM24} & \checkmark & ResNet-50 & 81.63 & 77.96 & 85.29 & 56.30 & 50.23 & 62.37 & 27.16 & 24.68 & 29.63 \\
QDFormer \cite{li2024qdformer} \textsubscript{CVPR24} & \ding{55} & ResNet-50 & 81.80 & 77.60 & 86.00 & 61.55 & 59.60 & 63.50 & $\ddagger$ & $\ddagger$ & $\ddagger$ \\
UFE \cite{liu2024audio} \textsubscript{CVPR24} & \ding{55} & ResNet-50 & 83.23 & 78.96 & 87.50 & 60.19 & 55.88 & 64.50 & $\ddagger$ & $\ddagger$ & $\ddagger$ \\
CAVP \cite{chen2024unraveling} \textsubscript{CVPR24} & \ding{55} & ResNet-50 & 83.84 & 78.78 & 88.89 & 61.48 & 55.82 & 67.14 & 32.83 & 30.37 & 35.29 \\
SEIM \cite{li2024selm} \textsubscript{ACM MM24} & \ding{55} & ResNet-50 & 81.40 & 76.60 & 86.20 & 60.05 & 54.50 & 65.60 & 34.55 & 31.90 & 37.20 \\
COMBO \cite{yang2024cooperation} \textsubscript{CVPR24} & \checkmark & ResNet-50 & 85.90 & \underline{81.70} & \underline{90.10} & 60.55 & 54.50 & 66.60 & 35.30 & 33.30 & 37.30 \\
CPM \cite{chen2025cpm} \textsubscript{ECCV24} & \ding{55} & ResNet-50 & \textbf{85.92} & 81.37 & \textbf{90.47} & \underline{65.40} & \underline{59.80} & \underline{71.00} & \underline{37.05} & \underline{34.53} & \underline{39.57} \\
\rowcolor{gray!20}
Ours & \checkmark & ResNet-50 & \underline{85.90} & \textbf{82.78} & 89.01 & \textbf{66.93} & \textbf{61.77} & \textbf{72.08} & \textbf{39.00} & \textbf{36.22} & \textbf{41.78} \\
\midrule 
AVSBench \cite{zhou2022audio} \textsubscript{ECCV22} & \ding{55} & PVT-v2 & 83.30 & 78.70 & 87.90 & 59.25 & 54.00 & 64.50 & 32.50 & 29.80 & 35.20 \\
DiffusionAVS \cite{mao2023contrastive}\textsubscript{arXiv23} & \ding{55} & PVT-v2 & 85.79 & 81.38 & 90.20 & 64.54 & 58.18 & 70.90 & $\ddagger$ & $\ddagger$ & $\ddagger$ \\
AVSBG \cite{hao2024improving} \textsubscript{AAAI24} & \ding{55} & PVT-v2 & 86.06 & 81.71 & 90.40 & 60.95 & 55.10 & 66.80 & $\ddagger$ & $\ddagger$ & $\ddagger$ \\
CATR \cite{li2023catr} \textsubscript{ACM MM23} & \ding{55} & PVT-v2 & 85.50 & 81.40 & 89.60 & 64.50 & 59.00 & 70.00 & 35.65 & 32.80 & 38.50 \\
AVSegFormer \cite{gao2024avsegformer} \textsubscript{AAAI24} & \ding{55} & PVT-v2 & 85.98 & 82.06 & 89.90 & 63.83 & 58.36 & 69.30 & 39.33 & 36.66 & 42.00 \\
UFE \cite{liu2024audio} \textsubscript{CVPR24} & \ding{55} & PVT-v2 & 86.78 & 83.15 & 90.40 & \underline{66.43} & \underline{61.95} & 70.90 & $\ddagger$ & $\ddagger$ & $\ddagger$ \\
SEIM \cite{li2024selm} \textsubscript{ACM MM24} & \ding{55} & PVT-v2 & 87.35 & 83.50 & 91.20 & 65.80 & 60.30 & \underline{71.30} & 44.10 & 41.30 & \underline{46.90} \\
COMBO \cite{yang2024cooperation} \textsubscript{CVPR24} & \checkmark & PVT-v2 & \underline{88.30} & \underline{84.70} & \underline{91.90} & 65.20 & 59.20 & 71.20 & \underline{44.10} & \underline{42.10} & 46.10 \\
\rowcolor{gray!20}
Ours & \checkmark & PVT-v2 & \textbf{90.07} & \textbf{86.63} & \textbf{93.51} & \textbf{69.89} & \textbf{64.38} & \textbf{75.39} & \textbf{48.16} & \textbf{45.03} & \textbf{51.28} \\
\bottomrule
\end{tabular}
}
\label{Quantitative comparison}
\end{table*}

\noindent\textbf{Datasets.} We conduct experiments on the AVS-Object \cite{zhou2022audio} and AVS-Semantic (AVSS) \cite{zhou2024audio} benchmarks. AVS-Object contains two subsets: single sound source segmentation (S4) and multiple sound source segmentation (M3), each with five sampled frames. The training/validation/testing sample capacities for S4 and M3 are 3452/740/740 and 296/64/64, respectively. Unlike AVS-Object, which only provides binary masks, AVSS further offers class labels and extends to ten sampled frames, comprising 12,356 (8498/1304/1554) videos across 70 categories.

\noindent\textbf{Implementation Details.} We leverage the PyTorch toolbox and execute our algorithm on NVIDIA A100 GPUs. All video frames are resized to 224×224. For the training phase, we employ the AdamW optimizer with the batch size of 8, the learning rate of 1e-4, and the total epochs of 60. During the testing phase, we do not apply any post-processing calibration, \textit{e.g.,} test-time augmentation \cite{shanmugam2021better}. For fair comparison, we adopt ResNet-50 \cite{he2016deep} and PVT-v2 \cite{wang2022pvt} pretrained on ImageNet as the visual backbone networks, VGGish \cite{hershey2017cnn} pretrained on Youtube-8M as the audio backbone. To generate implicit orthogonal text, we employ Latent Diffusion \cite{rombach2022high} to reduce computational complexity. Following \cite{yang2024cooperation}, we leverage  Multi-Scale Deformable Attention Transformer as the pixel decoder. The loss balancing hyper-parameter settings and more experiments are included in the supplementary material \footnote{Project: https://winter-flow.github.io/project/ICL}.

\noindent\textbf{Evaluation Metrics.} We utilize the Jaccard index $\mathcal{J}$, F-score $\mathcal{F}$, and mean value $\mathcal{J}\&\mathcal{F}$ as metrics for evaluation. Higher values indicate better model performance.

\begin{figure}[tb]
  \centering
  \includegraphics[width=0.48\textwidth]{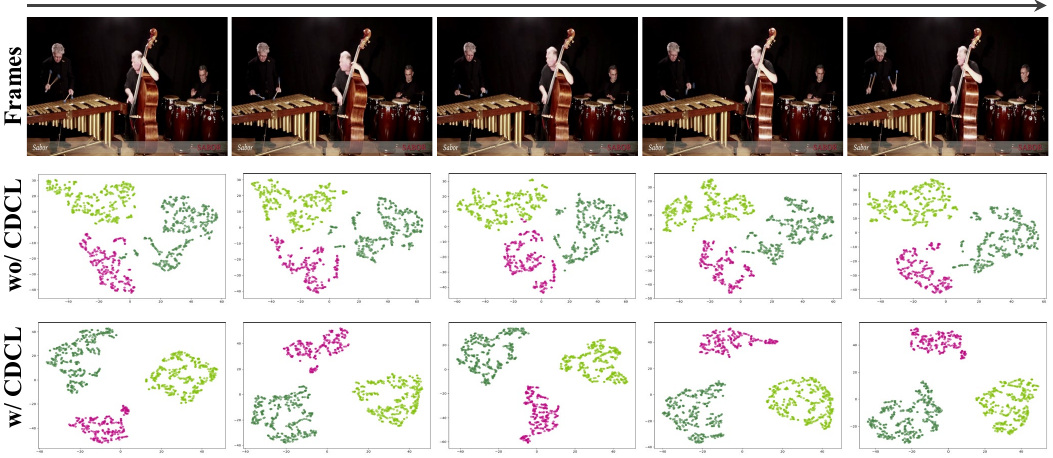}
  \caption{t-SNE visualization. 
    Drum (\textcolor{lightgreen}{\textbullet}), 
    Marimba (\textcolor{darkgreen}{\textbullet}), 
    Cello (\textcolor{mypurple}{\textbullet}).}
  \label{TSNE_vis}
\end{figure}

\begin{figure*}[tb]
  \centering
  \includegraphics[width=1\textwidth]{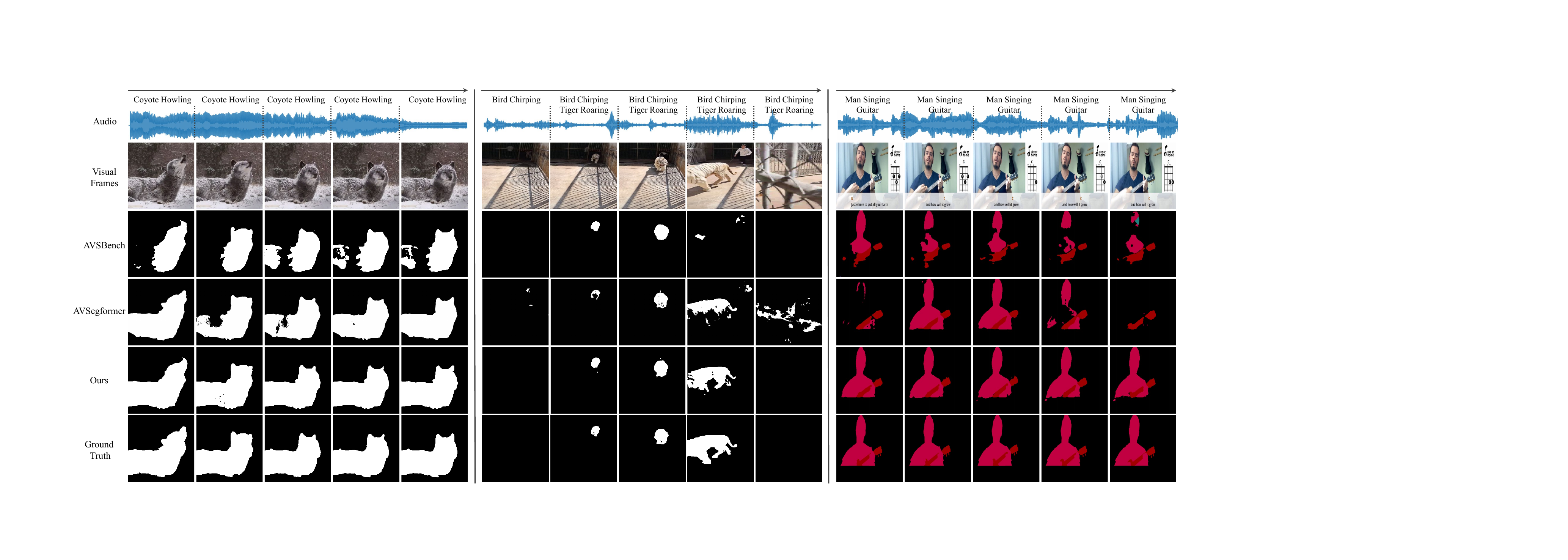}
  \caption{Qualitative comparison. From left to right, the samples are sourced from the S4, M3, and AVSS datasets. \textbf{Left:} The white snow on the coyote creates camouflage making it difficult to segment completely. \textbf{Middle:} Rapid movement and continuous noise interference (bird chirping); \textbf{Right:} Multiple audio and visual categories affected by subtitle background stripe.}
  \label{AVS_Vis_Comparision}
\end{figure*}

\begin{figure*}[tb]
  \centering
  \includegraphics[width=1\textwidth]{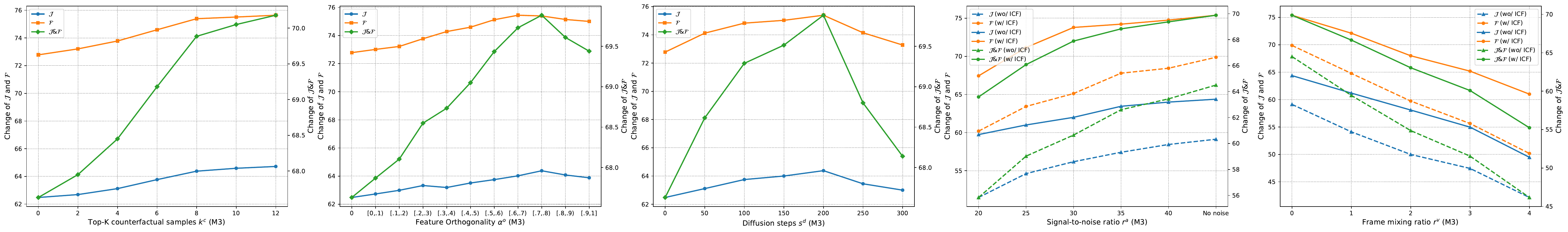}
  \caption{Quantitative ablation of Top-K $k^c$, orthogonality $\alpha^{o}$, number of diffusion steps $s^{d}$, signal-to-noise ratio $r^a$, and frame mixing $r^v$.}
  \label{Hyperparameter settings}
\end{figure*}

\subsection{Main Results}
\noindent\textbf{Quantitative Comparison.} In Table \ref{Quantitative comparison}, based on the visual backbone, \textit{i.e.,} ResNet-50 or PVT-v2, we categorize existing AVS methods into CNN-based and Transformer-based (partially overlapping). Our approach outperforms the second-best model, COMBO, by 1.77\%, 4.69\%, and 4.06\% in terms of $\mathcal{J}\&\mathcal{F}$ on three datasets. However, the superiority is not prominent on the S4 dataset. We analyze that the representation of visual scenes from single-source audio is relatively simple and pure, requiring less discriminative ability and representation space range. Conversely, in complex visual scenes with multiple-source audio, the lack of semantics may lead to catastrophic misalignment between audio and visual content. We focus on contrasting and analyzing two methods of incorporating text. BAVS \cite{liu2024bavs} introduces object category information, while our implicit text provides more comprehensive semantic understanding of video (including sequence and background) without prior knowledge. Compared to TeSO \cite{wang2024can} \footnote{Using Swin-Base \cite{liu2021swin} as the visual backbone network.}, which explicitly uses scene descriptions and relies on offline advanced LLMs (\textit{e.g.,} LLaMA2 \cite{touvron2023llama}) for sounding object perception, each step may result in incomplete and suboptimal outcomes. We instead directly match and fuse in the feature space, reducing manual intervention and noise introduction through end-to-end learning. Moreover, our approach outperforms DiffusionAVS \cite{mao2023contrastive}, which constructs latent diffusion model conditioned on audio and introduces paired contrastive learning, by an average of 4.82\%. We may attribute the performance gap to: 1) Inherent uncertainties (instabilities) in audio, \textit{i.e.,} semantic biases caused by multiple factors (\textit{e.g.,} noise disturbances); 2) Insufficient decoupling, lacking accurate reference and sample partition.

\noindent\textbf{Qualitative Comparison.} In Figure \ref{AVS_Vis_Comparision}, we present visualization comparison results under various settings. With increasing audio and visual complexity, other methods produce some false negatives or false positives. Our approach remains effective in establishing robust cross-modality correlations and capturing temporal changes, accurately segmenting sounding objects and determining categories.

\begin{table}[t]
\caption{Quantitative ablation of components and strategies.}
\centering
\normalsize
\resizebox{\linewidth}{!}{%
\begin{tabular}{lll|lll|lll}
\toprule
\multicolumn{3}{c|}{Component/Strategy} & \multicolumn{3}{c|}{M3} & \multicolumn{3}{c}{AVSS} \\
\cmidrule(lr){1-3} \cmidrule(lr){4-6} \cmidrule(lr){7-9}
~ & ~ & ~ & $\mathcal{J}\&\mathcal{F} \uparrow$ & $\mathcal{J} \uparrow$ & $\mathcal{F} \uparrow$ & $\mathcal{J}\&\mathcal{F} \uparrow$ & $\mathcal{J} \uparrow$ & $\mathcal{F} \uparrow$ \\ 
\midrule
MIT & SC & CDCL & 64.51 & 59.13 & 69.88 & 41.86 & 38.48 & 45.23 \\
\cmidrule(lr){1-3} 
$\checkmark$ & \ding{55} & \ding{55} & 66.64 & 61.44 & 71.83 & 44.10 & 40.83 & 47.37 \\
\ding{55} & \ding{55} & $\checkmark$ & 65.84 & 60.77 & 70.90 & 43.72 & 39.99 & 47.45 \\
$\checkmark$ & $\checkmark$ & \ding{55} & 68.03 & 62.94 & 73.11 & 45.66 & 42.55 & 48.76 \\
$\checkmark$ & \ding{55} & $\checkmark$ & 67.63 & 62.48 & 72.78 & 46.04 & 42.96 & 49.11 \\
\rowcolor{gray!20}
$\checkmark$ & $\checkmark$ & $\checkmark$ & \textbf{69.89} & \textbf{64.38} & \textbf{75.39} & \textbf{48.16} & \textbf{45.03} & \textbf{51.28} \\
\midrule
Frame & Video & Segment & 64.51 & 59.13 & 69.88 & 41.86 & 38.48 & 45.23 \\
\cmidrule(lr){1-3}
$\checkmark$ & \ding{55} & \ding{55} & 65.07 & 59.89 & 70.25 & 42.34 & 38.89 & 45.79 \\
$\checkmark$ & $\checkmark$ & \ding{55} & 65.68 & 60.43 & 70.92 & 43.11 & 39.77 & 46.45 \\
\rowcolor{gray!20}
$\checkmark$ & $\checkmark$ & $\checkmark$ & \textbf{66.64} & \textbf{61.44} & \textbf{71.83} & \textbf{44.10} & \textbf{40.83} & \textbf{47.37} \\
\midrule
Feature-level \textsuperscript{*} & Inter-sample & Intra-sample & 67.63 & 62.48 & 72.78 & 46.04 & 42.96 & 49.11 \\
\cmidrule(lr){1-3}
$\checkmark$ & \ding{55} & \ding{55} & 67.21 & 61.96 & 72.45 & 45.84 & 42.43 & 49.24 \\
\ding{55} & $\checkmark$ & \ding{55} & 68.87 & 63.52 & 74.22 & 47.33 & 44.47 & 50.19 \\
\ding{55} & \ding{55} & $\checkmark$ & 68.82 & 63.83 & 73.81 & 47.07 & 44.11 & 50.02 \\
\rowcolor{gray!20}
\ding{55} & $\checkmark$ & $\checkmark$ & \textbf{69.89} & \textbf{64.38} & \textbf{75.39} & \textbf{48.16} & \textbf{45.03} & \textbf{51.28} \\
\midrule
Discrete & Continuous & $\mathcal{L}_{\rm ortho}$ & 67.63 & 62.48 & 72.78 & 46.04 & 42.96 & 49.11 \\
\cmidrule(lr){1-3}
$\checkmark$ & \ding{55} & \ding{55} & 68.51 & 63.13 & 73.88 & 46.83 & 43.85 & 49.81 \\
\ding{55} & $\checkmark$ & \ding{55} & 69.18 & 63.86 & 74.49 & 47.31 & 44.28 & 50.33 \\
\rowcolor{gray!20}
\ding{55} & $\checkmark$ & $\checkmark$ & \textbf{69.89} & \textbf{64.38} & \textbf{75.39} & \textbf{48.16} & \textbf{45.03} & \textbf{51.28} \\
\midrule
Pair Swap & Audio Replacement & Text Revision & 67.63 & 62.48 & 72.78 & 46.04 & 42.96 & 49.11 \\
\cmidrule(lr){1-3}
$\checkmark$ & \ding{55} & \ding{55} & 68.16 & 62.89 & 73.42 & 46.56 & 43.71 & 49.40 \\
\ding{55} & $\checkmark$ & \ding{55} & 68.35 & 63.05 & 73.64 & 46.61 & 43.45 & 49.77 \\
\ding{55} & \ding{55} & $\checkmark$ & 68.44 & 63.31 & 73.56 & 46.91 & 43.95 & 49.86 \\
\midrule
$\mathcal{L}_{\rm v \leftrightarrow t}$ & $\mathcal{L}_{\rm a \leftrightarrow t}$ & $\mathcal{L}_{\rm a \leftrightarrow v}$ & 68.03 & 62.94 & 73.11 & 45.66 & 42.55 & 48.76 \\
\cmidrule(lr){1-3}
$\checkmark$ & \ding{55} & \ding{55} & 68.69 & 63.42 & 73.96 & 46.73 & 43.80 & 49.66 \\
$\checkmark$ & $\checkmark$ & \ding{55} & 69.32 & 63.85 & 74.78 & 47.24 & 44.29 & 50.19 \\
\rowcolor{gray!20}
$\checkmark$ & $\checkmark$ & $\checkmark$ & \textbf{69.89} & \textbf{64.38} & \textbf{75.39} & \textbf{48.16} & \textbf{45.03} & \textbf{51.28} \\
\midrule
Prototype & Feature & Distribution & 68.03 & 62.94 & 73.11 & 45.66 & 42.55 & 48.76 \\
\cmidrule(lr){1-3}
$\checkmark$ & \ding{55} & \ding{55} & 68.32 & 63.12 & 73.52 & 46.22 & 43.26 & 49.17 \\
\ding{55} & $\checkmark$ & \ding{55} & 68.59 & 63.19 & 73.99 & 46.90 & 43.79 & 50.01 \\
\rowcolor{gray!20}
\ding{55} & \ding{55} & $\checkmark$ & \textbf{69.89} & \textbf{64.38} & \textbf{75.39} & \textbf{48.16} & \textbf{45.03} & \textbf{51.28} \\
\bottomrule
\end{tabular}%
}
\label{Quantitative ablation of proposed components}
\end{table}

\begin{figure}[tb]
  \centering
  \includegraphics[width=0.48\textwidth]{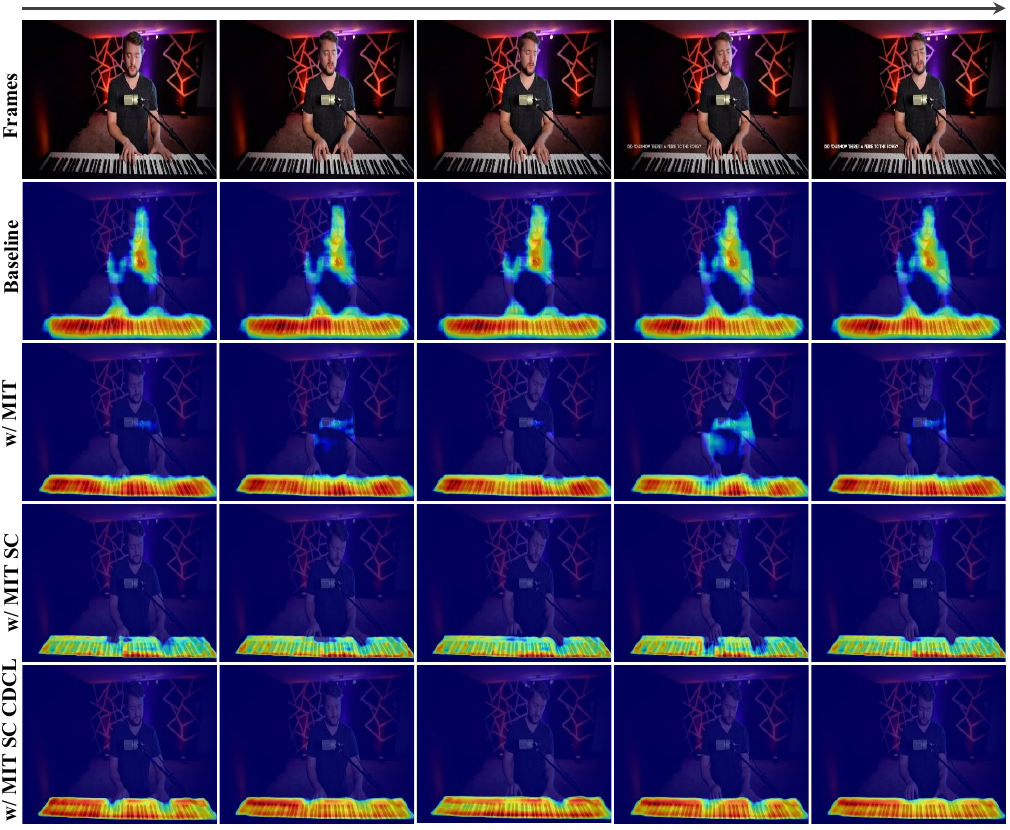}
  \caption{Qualitative ablation of proposed components. From top to bottom, the focus region transitions from salient object (person) to sounding object (piano) and is further enhanced by the CDCL.}
  \label{Qualitative ablation of proposed components}
\end{figure}


\begin{table}[t]
\caption{Effect of integrating the ICF into other approaches.}
\centering
\normalsize
\resizebox{\linewidth}{!}{%
\begin{tabular}{lll|lll|lll}
\toprule
\multicolumn{3}{c|}{Method} & \multicolumn{3}{c|}{M3} & \multicolumn{3}{c}{AVSS} \\
\cmidrule(lr){1-3} \cmidrule(lr){4-6} \cmidrule(lr){7-9}
~ & ~ & ~ & $\mathcal{J}\&\mathcal{F} \uparrow$ & $\mathcal{J} \uparrow$ & $\mathcal{F} \uparrow$ & $\mathcal{J}\&\mathcal{F} \uparrow$ & $\mathcal{J} \uparrow$ & $\mathcal{F} \uparrow$ \\ 
\midrule
AVSBench \cite{zhou2022audio} \textsubscript{ECCV22} & ~ & ~ & 59.25 & 54.00 & 64.50 & 32.50 & 29.80 & 35.20 \\
\rowcolor{gray!20}
\hspace{1.5em} w/ ICF & & ~ & \textbf{62.30} & \textbf{57.31} & \textbf{67.28} & \textbf{37.02} & \textbf{34.15} & \textbf{39.88}  \\
AVSegFormer \cite{gao2024avsegformer} \textsubscript{AAAI24} & ~ & ~ & 63.83 & 58.36 & 69.30 & 39.33 & 36.66 & 42.00 \\
\rowcolor{gray!20}
\hspace{1.5em} w/ ICF & ~ & ~ & \textbf{67.31} & \textbf{62.25} & \textbf{72.37} & \textbf{43.44} & \textbf{40.12} & \textbf{46.75} \\
\bottomrule
\end{tabular}%
}
\label{transition effect}
\end{table}

\subsection{Analysis and Discussion}
We analyze the effect of each component and framework (Table \ref{Quantitative ablation of proposed components}, Table \ref{transition effect} and Figure \ref{Qualitative ablation of proposed components}) and the settings of critical hyper-parameters  (Figure \ref{Hyperparameter settings}) on the M3 and AVSS datasets.

\noindent\textbf{Crucial Components.} The SC cannot work independently of the MIT. We integrate components through one-step and two-step processes. We find that: 1) Directly involving MIT-generated implicit text as an additional cue through cross-attention in modality interaction leads to insignificant performance gains; 2) When text information is not provided, the CDCL degenerates into visual-audio distribution contrast. As preceding components are gradually introduced, the gains increase. We analyze: 1) The implicit text is not purified and aligned, making carried uncertainties and heterogeneous features may disrupt the internal structure of other modalities, thereby reducing the discriminative representation of the shared space. Instead, with the assistance of the CDCL, textual features are implicitly injected, avoiding dedicated modules and suboptimal alignments. 2) With the addition of factual and counterfactual text, the expanded embedding space and accurate references promote modality cohesion and disentanglement.

\noindent\textbf{Multi-granularity Semantics.} Segment-level semantic understanding achieve the maximum gain, based on integrating video-level and frame-level. This may be attributed to segments balancing short-term and continuous contexts. Long sequence modeling provides global optimization directions, while intra-frame modeling offers scene spatial correlations, achieving complementarity.

\noindent\textbf{Counterfactual Dimensions, Space and Constraint.} The synergistic counterfactual effect within and between samples is greater than single one. We employ VQVAE \cite{van2017neural} to replace Diffusion to generate orthogonal representations in the discrete space, yet the performance is not promising. We attribute to the the potential decrease in semantic coherence and sample diversity when forcibly partitioning features into several groups. Conversely, continuous space is more conducive to ensuring the reasonableness and varying degrees. $\mathcal{L}_{\rm ortho}$ further controls deviations in the denoising.

\noindent\textbf{Counterfactual Strategies.} Inspired by \cite{jiang2024counterfactually}, we randomly swap half of the video and audio samples with other samples in the datasets to recombine and construct counterfactual pairs, \textit{i.e, Pair Swap}. Similar to \cite{singh2024looking}, we obtain sounding categories and utilize Text-To-Audio model \cite{huang2023make} to generate audio with different spectra for replacement, \textit{i.e., Audio Replacement}. We further leverage Qwen2-VL \cite{wang2024qwen2} to generate descriptions of videos and replace nouns to generate counterfactual explicit text, \textit{i.e., Text Revision}. The gain of \textit{Text Revision} is the largest, while \textit{Pair Swap} is the smallest. We analyze that: 1) Exchanging samples from the dataset alters the sequential structure but does not fundamentally change the representation space; 2) Clues generated by external models expand the representation space, while text enriches semantic variations. The above strategies require significant computational and storage expenses, and are difficult to dynamically adjust based on input. Our method employ online end-to-end training to adaptively equip counterfactual at different positions.

\noindent\textbf{Modality Contrastive Loss.} The tri-modal contrastive loss exhibits complementarity. We rely on masked average pooling or mean to generate prototypes to depict holistic information. Performance averages 1.76\% lower than distribution-based approach. We argue that distribution mining captures the general statistical probabilities of representations, avoiding anomalies and unstable interferences. However, treating each element indiscriminately with only the mean dilutes the correct representations. In Figure \ref{TSNE_vis},  we illustrate the transition from loose crossover to clustered separation before and after using the CDCL.

\noindent\textbf{Critical Hyper-parameters.} To balance performance and cost, we set $k^c=8$, $\alpha^{o}=[0.7,0.8)$, $s^d=200$ (the larger $s^d$, the closer the features approach pure noise). As $r^a$ decreases and $r^v$ (1/4 of M3) increases, the performance gap becomes more significant, with about a 10\% gain of  $\mathcal{J}\&\mathcal{F}$.

\noindent\textbf{Limitation.} We achieve matrix heterogeneity by Gram-Schmidt with the complexity of $\mathcal{O}(n^2)$. Besides, numerous crucial hyper-parameters require iterative experiments to determine the optimal combination. In the future, we aim to reduce the complexity (\textit{e.g.,} SVD \cite{klema1980singular}) and explore automated parameter selection methods (\textit{e.g.,} MoE \cite{jacobs1991adaptive}). 
\section{Conclusion}
\label{sec:conclusion}

Based on causal inference and semantic guidance, we aim to eliminate modality biases and model complex spatiotemporal patterns. Unlike explicitly generating textual descriptions and modifying attributes, we construct counterfactual on the implicitly generated text from the foundation model leveraging \textit{`if not...then...'}. The strategy prevents the model from relying on statistical regularities to form spurious associations. By orthogonalizing the latent space and introducing constraint factors to ensure controllable counterfactual generation, we further establish decoupling and coherence among modalities in the Gaussian space. Our ICF is easily integrated into other frameworks.

\vspace{0.1cm}
\noindent \textbf{Acknowledgements:} This work was supported in part by the National Natural Science Foundation of China under grant U23B2011, 62102069, U20B2063 and 62220106008, the Key R\&D Program of Zhejiang under grant 2024SSYS0091, the Sichuan Science and Technology Program under Grant 2024NSFTD0034.

{
    \small
    \bibliographystyle{ieeenat_fullname}

}

\end{document}